# Face Alignment Using Active Shape Model And Support Vector Machine

**Le Hoang Thai**                                    lhthai@fit.hcmus.edu.vn
*Department of Computer Science*
*University of Science*
*Hochiminh City, 70000, VIETNAM*

**Vo Nhat Truong**                                   vntruong@gmail.com
*Faculty/Department/Division*
*University of Science*
*Hochiminh City, 70000, VIETNAM*

## Abstract

The Active Shape Model (ASM) is one of the most popular local texture models for face alignment. It applies in many fields such as locating facial features in the image, face synthesis, etc. However, the experimental results show that the accuracy of the classical ASM for some applications is not high. This paper suggests some improvements on the classical ASM to increase the performance of the model in the application: face alignment. Four of our major improvements include: i) building a model combining Sobel filter and the 2-D profile in searching face in image; ii) applying Canny algorithm for the enhancement edge on image; iii) Support Vector Machine (SVM) is used to classify landmarks on face, in order to determine exactly location of these landmarks support for ASM; iv) automatically adjust 2-D profile in the multi-level model based on the size of the input image. The experimental results on Caltech face database and Technical University of Denmark database (imm_face) show that our proposed improvement leads to far better performance.

**Keywords:** Face Alignment, Active Shape Model, Principal Component Analysis.

## 1. INTRODUCTION

Face recognition is the problem to search human faces in large image database. In detail, a face recognition system with the input of an arbitrary image will search in database to output people's identification in the input image. The face recognition system's stages are illustrated in Figure 1 [5].

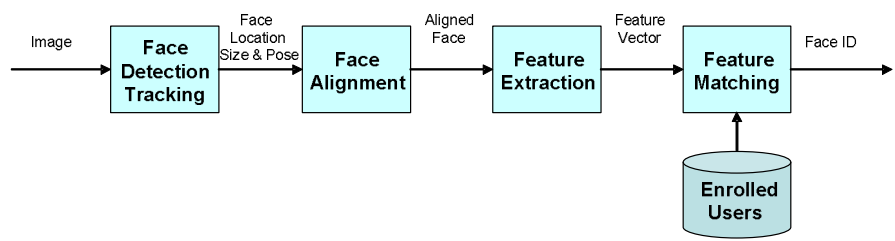

**FIGURE 1:** Structure of a face recognition system.

The face alignment is one of important stages of the face recognition. Moreover, face alignment is also applied for other face processing applications; such as face modeling and synthesis. Its objective is to localize the feature points on face images such as the contour points of eye, nose, mouth and face (illustrated in Figure 2).

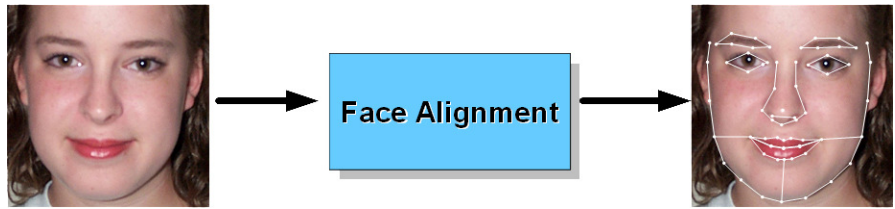

**FIGURE 2:** Face alignment.

There are many face alignment methods. Two popular face alignment methods are Active Shape Model (ASM) [16] and Active Appearance Model (AAM)[8] are proposed by Cootes et al.

The two methods use a statistical model to parameterize a face shape with Principal Component Analysis (PCA) method. However, their feature model and optimization are different. ASM algorithm has a 2-stage loop: in the first stage, given the initial labels, searching for a new position for every label point in its local region which best fits the corresponding local 1-D profile texture model; in the second stage, updating the shape parameters which best fits these new label positions. AAM method uses its global appearance model to directly conduct the optimization of shape parameters. Owing to the different optimization criteria, ASM performs more precisely on shape localization, and is quite more robust to illumination and bad initialization. In the paper extent, we develop the classical ASM method to create a new method named ASM-SVM which has achieved better results.

Because ASM only uses a 1-D profile texture feature, which is not enough to distinguish feature points from their local regions, the ASM algorithm often fall into local minima problem in the local searching stage. A few representative texture features and pattern recognition methods are proposed to reinforce the ASM local searching, e.g. Gabor wavelet, Haar wavelet, Ranking-Boost, Fisher Boost and MLP-ASM (Perceptron network) [5]. Nevertheless, an accurate local texture model to large data sets is still unachieved target.

In this paper, we propose the improvements in the local search of ASM. The main improvements are followed: first, build a model combining Sobel filter and the 2-D profile in searching face in image; second, applying Canny algorithms for the enhancement edge in image; third, support vector machine (SVM) is used to classify landmarks on face, in order to determine the exact location of these landmarks support for ASM; last, automatically adjust 2-D profile in the multi-level model based on the size of the input image.

The paper is structured as follows: Section 2, we present the classical ASM algorithm, section 3 presents details of our improvements and Section 4 presents experimental results and Section 5 presents conclusion and future works.

## 2. CLASSICAL ASM ALGORITHM

### 2.1 Training stage

A shape can be represented by n points $\{(x_i, y_i)\}$ as a 2n-D element vector, $X = (x_1, y_1, \ldots, x_n, y_n)^T$. With training shape S ($S = \{X_i\}$), we perform statistical shape on the same coordinates.

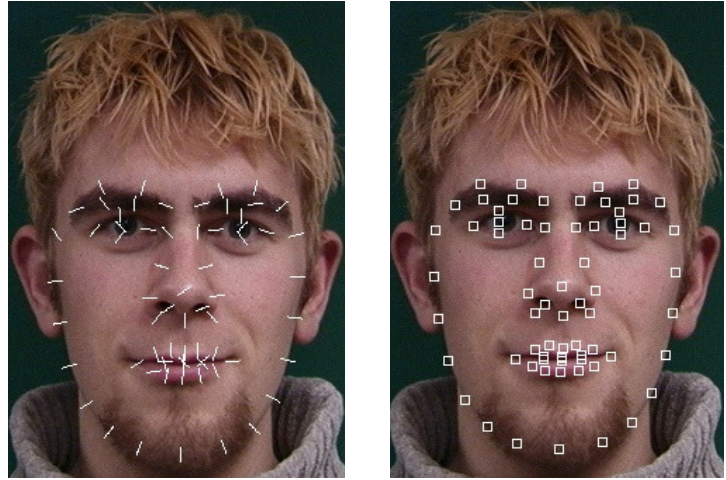

**FIGURE 3:** Local features to be built in the period of training (a) typical 1-D (b) typical 2-D

The shape of the training set S are aligned by algorithms Generalized Procrustes Analysis (GPA) [12]. Average shape ($\overline{x}$) is the average shape vector of all alignment shape. PCA is applied to calculate this shape and the covariance matrix is chosen so that accounted for 97.5% of the total value of training set that are arranged from large to small and used to store as the corresponding eigenvector matrix (P).

In next stage, we determine the gray level to create the statistical model of gray level around the landmarks to build a subspace represents the change of training shape. 1-D profiles are constructed by the gray level of points on the line with fixed length. These straight lines are orthogonal to the edge of this shape at the landmark. The gray level sample is stored as a vector that is then standardized by replacing each element of the vector with the intensity of gray levels (the difference of gray level at that point and the preceding point) and then dividing the magnitude of the vector average. Average profile (of all files in training set) is called average profile vector ($\overline{g}$) and covariance matrix of all the vector present as $S_g$. Average profile vector and covariance matrix are generated for each point and three level of the pyramid model (each image in the pyramid is half the image size of it before).

Similar, training data can be calculated by 2-D profile that is created at each landmark by the derivation of gray level image (the sum of square derivation in x and y directions) .Result matrix is transformed into a vector and then normalized by the sigmoid transformation for each specific element of profile $g'_i$ as follow equation:

$$g'_i = \frac{g_i}{(\left|g_i\right| + q)} \qquad (1)$$

q: const.

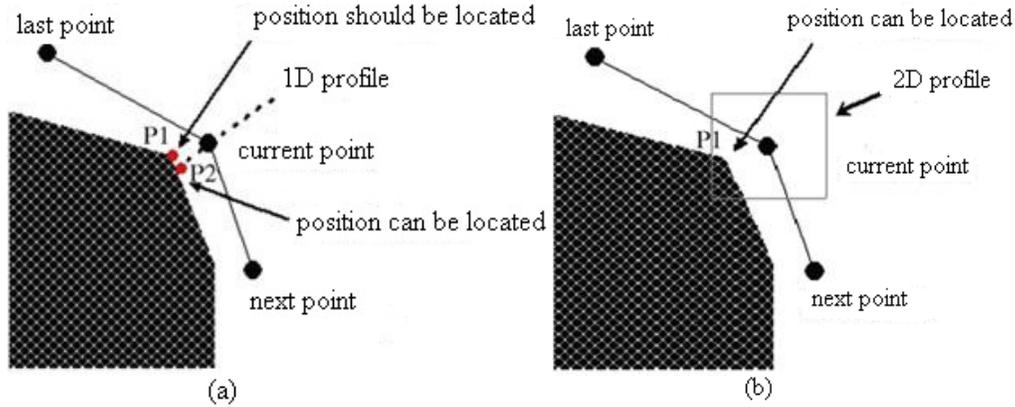

**FIGURE 4:** Illustrate local features: (a) typical 1D and (b) typical 2D

Using 1-D profile to find landmark in some cases is not accurate. For example, Figure 4(a) illustrates the case that desired position is P1. However, 1-D profile achieves P2 point instead of P1. Hence, 2-D profile is necessary to solve this problem (Figure 3 (b)). The desired location is P1 can be determined exactly by 2-D profile. Moreover, misplaced errors will reduce by using 2-D profile.

### 2.2 Alignment stage

In alignment phase, faces in test images will be identified in first step. Face detection algorithm, Viola Jones in OpenCV [11], is used for this step. After detecting the location of faces in images, similar transformations (scale, rotation and translation) will operate on the shape model that represent the face (constructing from the training data set) to fit this model to test face (the face is detected by OpenCV). Achieving shape will use as initial shape. A loop on the initial shape would be made to find suitable final landmark. These landmarks will form a shape that best suits the considered face image.

Typical multi-level model is built for image at each level by the method as in training phase. The process of identifying profile start from the lowest level of the pyramid (level 2) and gradually move up to the highest level (level 0) (Figure 4). Fluctuations of the landmarks are highest at the lowest level and they are smaller at higher levels. Best location of the landmark is determined by establishing the profile of the candidate neighboring around the landmarks. The candidate points that have nearly all features of average landmark will be selected as the new location of landmark. The weighting function that use in ASM to determine at this landmark is the smallest Mahalanobis distance ($f_1(g)$) of candidates (g) with average profile $\overline{g}$ by the following equation:

$$f_1(x) = (g - \overline{g})^T S_g^{-1} (g - \overline{g}) \qquad (2)$$

Searching process on the 2-D profile with size 15x15 pixels around the landmark will operate at all levels of multi-level model.

When all landmarks move to the best location, a new shape ($x_l$) needs to be converted into an appropriate shape and to represent the boundary of face. This is done by Equation (3).

$$x_L = \overline{x} + Pb \qquad (3)$$

$x_L$: the closest shape vector ($x_l$)
$\overline{x}$ : average shape
P: eigenvector matrix

b: coefficient vector that is predicted to generate the face shape.

b is calculated by performing a loop so that the distance of formula (4) is the smallest.

$$dist(x_I, T(\overline{x} + Pb)) \tag{4}$$

T is a transformation that makes minimizing distance between $x_I$ and $\overline{x} + Pb$. [9] represents the algorithm that finds b and T. When we get vector b, $b_i$ is $i^{th}$ element of vector b and it have to be between $-\alpha\sqrt{\lambda_i}$ and $+\alpha\sqrt{\lambda_i}$ with $\alpha$ is 3 and $\lambda_i$ is $i^{th}$ eigenvalue.

The limitation of these values can ensure that the generated shape is similar to those in the original training set.

At each level of multi-level model, a loop will be done until convergence (no significant change of landmark position in two consecutive loops). If convergence is done at lowest level, the shape will be changed by scale transformation and used as initial position for the next level of multi-level model. This process continues until convergence and achieving final landmarks at the highest level of multi-level model.

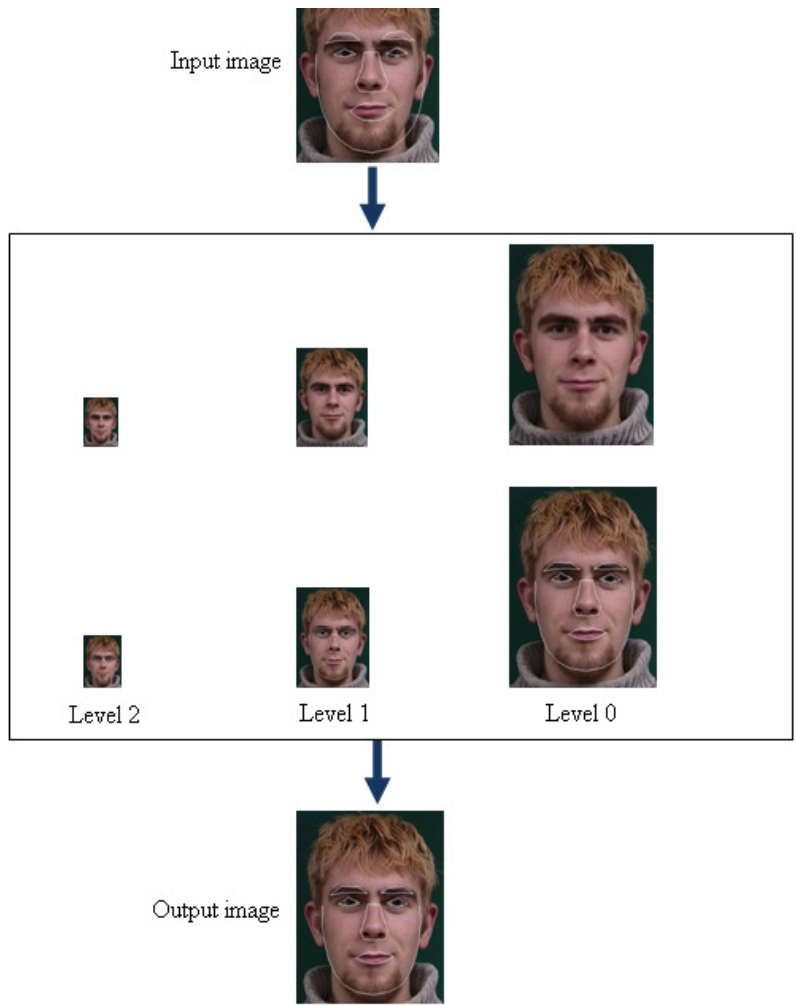

**FIGURE 5:** Illustrate alignment of multi-level model

## 3. IMPROVEMENT FOR ASM

### 3.1 Combining 2-D profile and Sobel filter

In order to balance brightness in image as well as distinguish between high and low frequency variations in image. In this paper, we determine the 2-D profile for each point by combining histogram equalization and Sobel filter as follow:

Step 1: Using the Histogram balancing algorithm to normalize brightness of image.
Step 2: Using the Sobel filter function in two directions x, y. Constructing texture matrix, with value of each point in the matrix is the square root of the sum of squared derivative in two directions (x and y).
Step 3: Normalizing result matrix to a vector by formula (5).

$$g'_i = \frac{g_i}{\sum g_i} \qquad (5)$$

### 3.2 Enhancement edge by Canny algorithm

To increase more accurately for fitting points that lie along the boundary, we use the weighting function (6).

$$f_2(g) = (c - \mathrm{I})(g - \overline{g})^T S_g^{-1}(g - \overline{g}) \qquad (6)$$

In the function $f_2(g)$, $\mathrm{I}$ is the gray value at candidate point and has value 0 (for the point not on the boundary) or 1 (for the point on the boundary). $\mathrm{I}$ determined based on enhancement edge by Canny algorithm [11]. c is a constant and we choose 2 for our experiments. Function $f_2(g)$ can increase the ability to searching landmark on the boundary of shape that hard to find in classical algorithm.

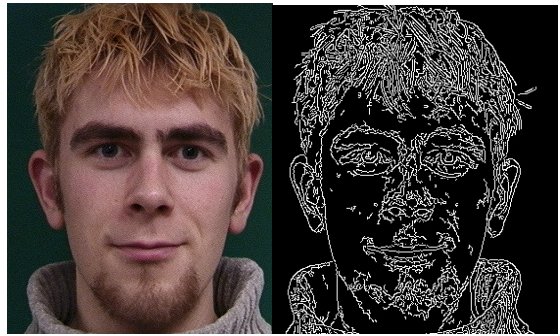

**FIGURE 6:** Illustrates the edge detection algorithm (Canny): (a) original image (b) resulting image

### 3.3 Applying SVM to find landmarks

In the classical ASM method, the PCA does not consider the distinction between the positive sample (the points represent the model (Section 4)) and negative sample (the points are not positive sample). So the searching landmark process often falls into local extreme values. To distinguish between positive sample and negative sample, we chose SVM method [10] because this method generalizes learning sample (without learning much data as other classification methods) and minimizes the structure error that increases the classified ability. In this paper, we use linear SVM.

For each point, we determine local 2-D profile with this. Next, the positive samples (the landmark) are selected from the focal point, whereas the negative samples (the point is not the landmark) randomly select window which has same size and different focal point with positive sample. Algorithms search candidate around the current landmark to determine the new landmark:

**Input**: shape X {(x$_i$, y$_i$)}
**Output**: shape X'{(x'$_i$, y'$_i$)}
For each point (x$_i$, y$_i$) of X
For each window has focus point (x', y') that belong to window with focus point (x$_i$, y$_i$).
Applying SVM in the window (x', y'). If the return value is +1, the point (x', y') lies on the boundary, otherwise returns -1.
Selecting a point (x ', y') that value of function f$_2$(g) is the smallest. This point is the new landmark.

### 3.4 Adjusting profile length

In the classical ASM algorithm, the lengths of profile windows are the same size at each level in multi-level model. However, from experiment, we found that the shift of the points in each level is different. At higher levels, the shift is smaller. Moreover, the shift is very small when the candidates are close to destination. From this observation, we adjust profile length according to different levels. Adjusting profile length save the computational cost and increase the accuracy of landmark determination. Length of the window is our proposal to reduce the level of ascent. The length of the first level is L, L/2 for level 2, and L/4 for the final level. In this experiment we use a length for the first level is 15 pixels.

## 4. EXPERIMENTAL RESULTS

Face shape are made from 68 landmarks that extract with specific groups as follow: face boundary (15 points), right eyebrow (8 points), left eyebrow (8 points), left eye (8 points), right eye (8 points), nose (9 points) and oral (12 points).

Evaluating performance of our proposed method and other methods, we use the average error calculation function as follow:

$$E_{ave} = \frac{1}{n} \times \frac{1}{k} \sum_{i=1}^{n} \sum_{j=1}^{k} abs(x(i,j) - pos(i,j)) \qquad (7)$$

### 4.1 CalTech database

Caltech face image data includes 450 jpeg images with 896 x 592 sizes of 27 subjects.
We randomly select 300 images for training and the remaining 150 images for test. Table 1 illustrates the test results.

| Method | Average Error (E$_{ave}$) |
|---------|---------|
| ASM | 14.021 |
| MLP-ASM | 12.403 |
| ASM-SVM | 10.548 |

**TABLE 1**: Experimental results on the Caltech database

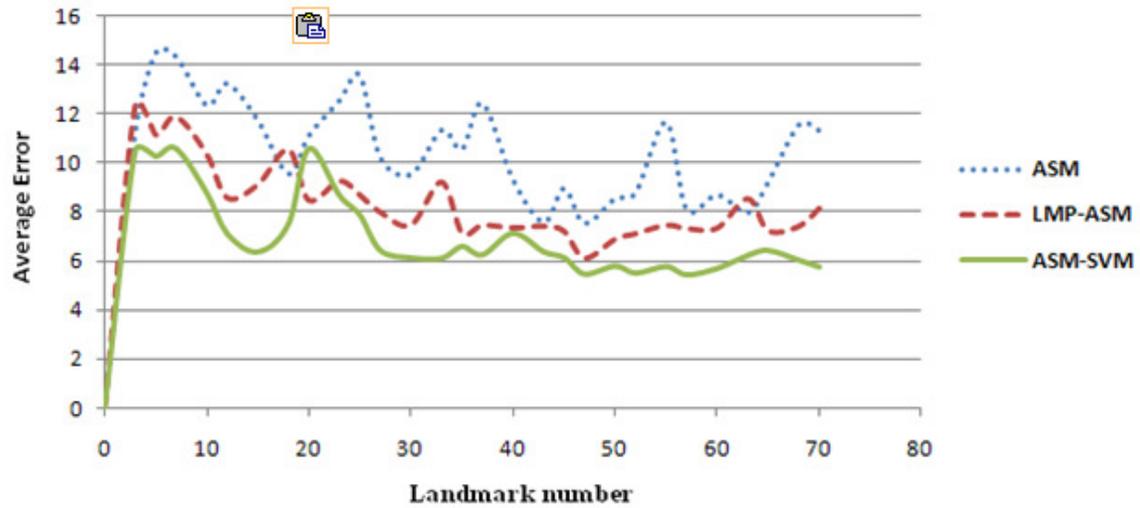

**FIGURE 7:** Comparison between the classical ASM, MLP-ASM and ASM-SVM.

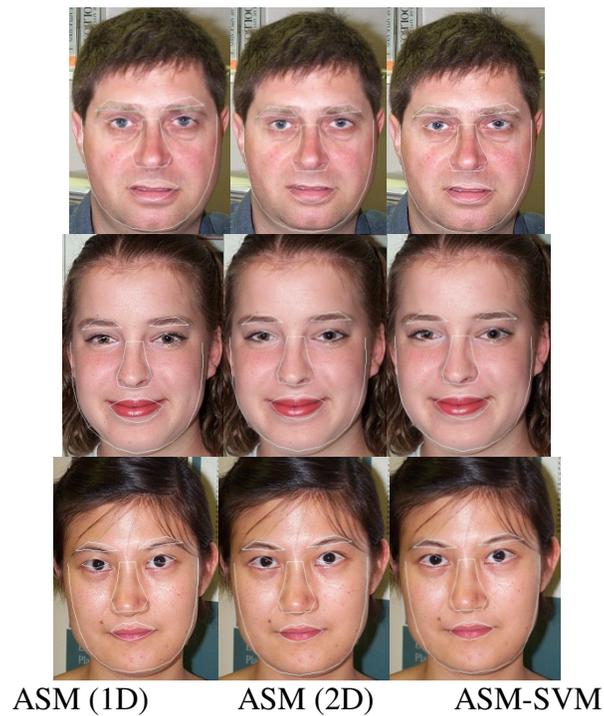

ASM (1D)    ASM (2D)    ASM-SVM

**FIGURE 8:** Some experimental results

## 4.2   DTU database

DTU face image data includes 240 jpeg images with 640 x 480 sizes of 40 subjects.

We randomly select 160 images for training and the remaining 80 images for test. Table 2 illustrates the test results.

| Method | Average Error ($E_{ave}$) |
|---|---|
| ASM | 11.751 |
| MLP-ASM | 9.147 |
| ASM-SVM | 7.176 |

**TABLE 2**: Experimental results on the DTU database

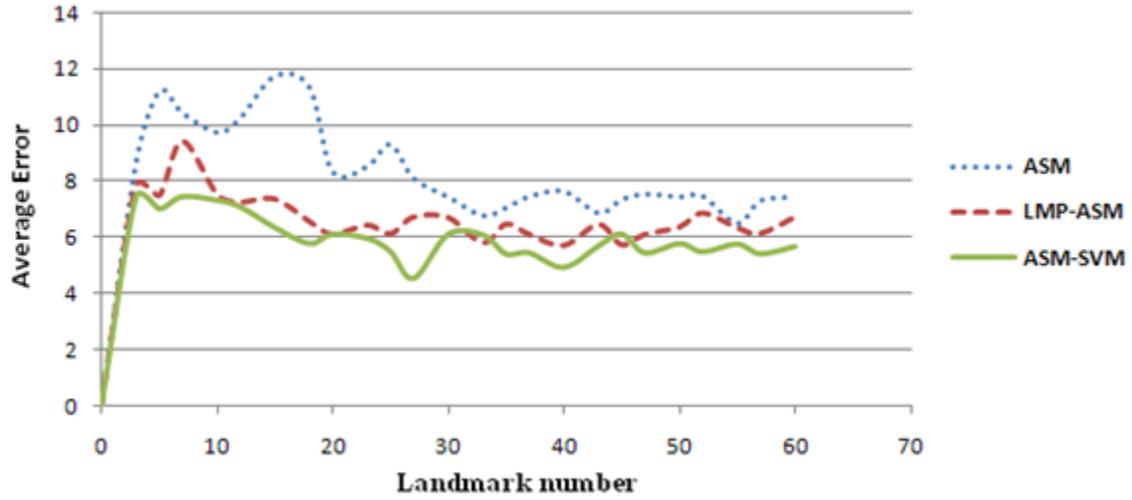

**FIGURE 9:** Comparison between the classical ASM, MLP-ASM and ASM-SVM.

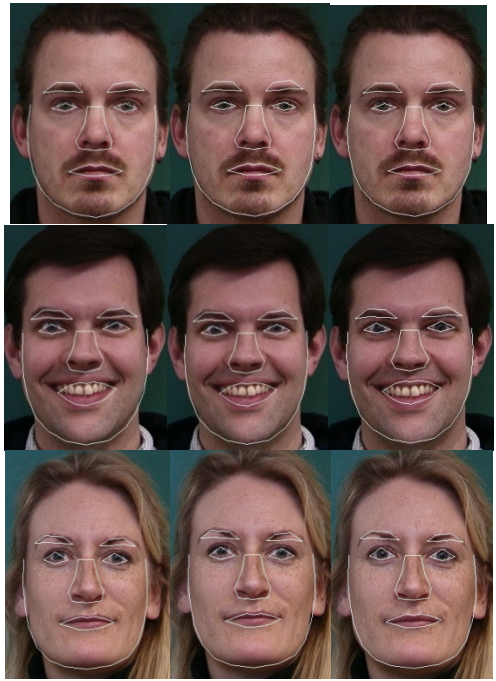

**FIGURE 10:** Some experimental results

**4.3 Others**

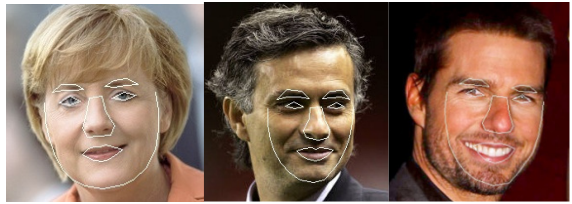

**FIGURE 11:** Results on real images obtained from the internet.

Reviews: with experimental results on two databases: Caltech and DTU, our approach is general and can be applied to many different databases. With using edge detection methods, our method gets high efficiency when compare to classical ASM at landmarks on the boundary of face.

# 5. CONSLUSION & FUTURE WORKS

In this paper, we propose an alignment model using ASM combine with SVM. Instead of using 1-D profile model, we use 2-D profile model and combine with Sobel filter function to new landmarks that are neighbored with original ones. This model is useful for finding landmarks, which bases on the strong classifier of SVM and the distance measuring of Mahalanobis, as well as determine strong edges to increase the accuracy of determining landmarks. General and powerful classifier of proposed model makes ASM more efficient. The result of the comparison proposed method to classical ASM, bases on Caltech database and DTU database (imm_face), show that our proposed improvements are better performance.

In the future, we can use hierarchical approach. Firstly, using ASM for the global features (face boundary), and then we use ASM for the local features (left eye, right eye, nose, mouth). Combining this method with Expectation Maximization Algorithm is useful for adjusting incorrect landmarks.